%% file: _main.tex
\input{_constants}
\arxiv 

\pdfoutput=1
\documentclass[10pt,twocolumn,letterpaper]{article}
\input{cvpr_header}
\usepackage[accsupp]{axessibility}
\begin{document}
\title{\paperTitle}
\author{\authorBlock}
\maketitle

\input{00_abstract}
\input{01_intro}
\input{02_related}

\input{03_preliminary}
\input{04_method}

\input{05_experiments}

\input{10_conclusion}

\xhdr{Acknowledgements}
This work was supported in part by the National Key Research and Development Program of China No. 2020AAA0106300, National Natural Science Foundation of China (No. 62250008, 62222209, 62102222, 61936011, 62206149), Beijing National Research Center for Information Science and Technology (BNRist) under Grant No. BNR2023RC01003, Beijing Key Lab of Networked Multimedia, China National Postdoctoral Program for Innovative Talents No. BX20220185, and China Postdoctoral Science Foundation No. 2022M711813.

{\small
\bibliographystyle{ieee_fullname}
\bibliography{11_references}
}

\ifarxiv \clearpage 
\input{12_appendix}


\end{document}

%% file: _constants.tex
\def\cvprPaperID{5751}
\def\confName{CVPR}
\def\confYear{2023}

\def\paperTitle{Adversarially Robust Neural Architecture Search for Graph Neural Networks}

\def\authorBlock{
    Beini Xie\textsuperscript{1}\thanks{Equal contribution: {\tt \{xbn20,changh17\}@mails.tsinghua.edu.cn }} \qquad
    Heng Chang\textsuperscript{1}\footnotemark[1] \qquad
    Ziwei Zhang\textsuperscript{1} \qquad 
    Xin Wang\textsuperscript{1}\footnotemark[2] \qquad 
    Daixin Wang\textsuperscript{2} \qquad \\
    Zhiqiang Zhang\textsuperscript{2} \qquad
    Rex Ying\textsuperscript{3} \qquad
    Wenwu Zhu\textsuperscript{1}\thanks{Corresponding authors: {\tt \{xin\_wang,wwzhu\}@tsinghua.edu.cn }} \\
    \textsuperscript{1}Tsinghua University \qquad
    \textsuperscript{2}Ant Group \qquad
    \textsuperscript{3}Yale University \\
    
}

\newif\ifreview 
\newif\ifarxiv \newcommand{\arxiv}{\arxivtrue}
\newif\ifcamera 
\newif\ifrebuttal 

%% file: cvpr_header.tex
\ifreview \usepackage[review]{cvpr} \fi
\ifarxiv \usepackage[pagenumbers]{cvpr} \fi
\ifrebuttal \usepackage[rebuttal]{cvpr} \fi
\ifcamera \usepackage{cvpr} \fi

\usepackage{graphicx}
\usepackage{amsmath}
\usepackage{amssymb}
\usepackage{booktabs}
\usepackage{mathtools}

\input{_macros}  

\usepackage{xr-hyper}

\makeatletter
\newcommand*{\addFileDependency}[1]{
  \typeout{(#1)}
  \@addtofilelist{#1}
  \IfFileExists{#1}{}{\typeout{No file #1.}}
}

\makeatother

\usepackage[pagebackref,breaklinks,colorlinks]{hyperref}
\usepackage[capitalize]{cleveref}
\crefname{section}{Sec.}{Secs.}
\crefname{table}{Table}{Tables}
\crefname{figure}{Fig.}{Figs.}

%% file: _macros.tex
\usepackage{times}
\usepackage{microtype}
\usepackage{epsfig}
\usepackage[table,xcdraw]{xcolor}
\usepackage{caption}
\usepackage{float}
\usepackage{placeins}
\usepackage{color, colortbl}
\usepackage{stfloats}
\usepackage{enumitem}
\usepackage{tabularx}
\usepackage{xstring}
\usepackage{multirow}
\usepackage{xspace}
\usepackage{url}
\usepackage{xcolor}
\usepackage[hang,flushmargin]{footmisc}

\usepackage{amsfonts}

\usepackage{wrapfig}
\usepackage{bm}
\usepackage[noend,ruled,linesnumbered]{algorithm2e}
\usepackage{graphicx}
\usepackage{mathtools}

\ifcamera \usepackage[accsupp]{axessibility} \fi

% ...

\newcommand{\xhdr}[1]{\noindent{{\bf #1.}}}

\ifarxiv  \fi
\newcommand{\revision}[1]{{{\textcolor{black}{#1}}}}

\newcommand{\R}[1]{{%
    \textbf{%
        \ifstrequal{#1}{1}{\textcolor{red}{R#1}}{%
        \ifstrequal{#1}{2}{\textcolor{blue}{R#1}}{%
        \ifstrequal{#1}{3}{\textcolor{magenta}{R#1}}{%
        \ifstrequal{#1}{4}{\textcolor{teal}{R#1}}{%
                           \textcolor{cyan}{R#1}%
        }}}}%
    }%
}}

%% file: 00_abstract.tex
\begin{abstract}
Graph Neural Networks (GNNs) obtain tremendous success in modeling relational data. Still, they are prone to adversarial attacks, which are massive threats to applying GNNs to risk-sensitive domains. 
Existing defensive methods neither guarantee performance facing new data/tasks or adversarial attacks nor provide insights to understand GNN robustness from an architectural perspective. 
Neural Architecture Search (NAS) has the potential to solve this problem by automating GNN architecture designs. Nevertheless, current graph NAS approaches lack robust design and are vulnerable to adversarial attacks. To tackle these challenges, we propose a novel Robust Neural Architecture search framework for GNNs (G-RNA). Specifically, we design a robust search space for the message-passing mechanism by adding graph structure mask operations into the search space, which comprises various defensive operation candidates and allows us to search for defensive GNNs. Furthermore, we define a robustness metric to guide the search procedure, which helps to filter robust architectures. In this way, G-RNA helps understand GNN robustness from an architectural perspective
and effectively searches for optimal adversarial robust GNNs. 
Extensive experimental results on benchmark datasets show that G-RNA significantly outperforms manually designed robust GNNs and vanilla graph NAS baselines by 12.1\% to 23.4\% under adversarial attacks.
\end{abstract}

%% file: 01_intro.tex
\section{Introduction}
\label{sec:intro}

Graph Neural Networks are well-known for modeling relational data and are applied to various downstream real-world applications like recommender systems~\cite{wang2019heterogeneous}, knowledge graph completion~\cite{lin2015learning}, traffic forecasting~\cite{diao2019dynamic}, drug production~\cite{yu2022structure}, etc. Meanwhile, like many other deep neural networks, GNNs are notorious for their vulnerability under adversarial attacks~\cite{sun2018adversarial}, especially in risk-sensitive domains, such as finance and healthcare. 
Since GNNs model node representations by aggregating the neighborhood information, an attacker could perform attacks by perturbing node features and manipulating relations among nodes~\cite{KDD2018Adversarial}.
For example, in the user-user interaction graph, a fraudster may deliberately interact with other important/fake users to mislead the recommender system or fool credit scoring models~\cite{wang2019semi}.

A series of defense methods on graph data have been developed to reduce the harm of adversarial attacks. 
Preprocessing-based approaches like GCN-SVD~\cite{entezari2020all} and GCN-Jaccard~\cite{wu2019adversarial} conduct structure cleaning before training GNNs, while attention-based models like RGCN~\cite{RGCN} and GNN-Guard~\cite{zhang2020gnnguard} learn to focus less on potential perturbed edges. 
However, these methods rely on prior knowledge of the attacker. For example, GCN-SVD leverages the high-rank tendency of graph structure after Nettack~\cite{zugner2018adversarial}, and GCN-Jaccard depends on the homophily assumption on the graph structure.
As a result, current approaches may fail to adapt to scenarios when encountering new data and tasks or when new attack methods are proposed.
Additionally, previous methods largely neglect the role of GNN architectures in defending against adversarial attacks, lacking an architectural perspective in understanding GNN robustness.

\begin{figure*}[htbp]
    \centering
    \includegraphics[width=\textwidth]{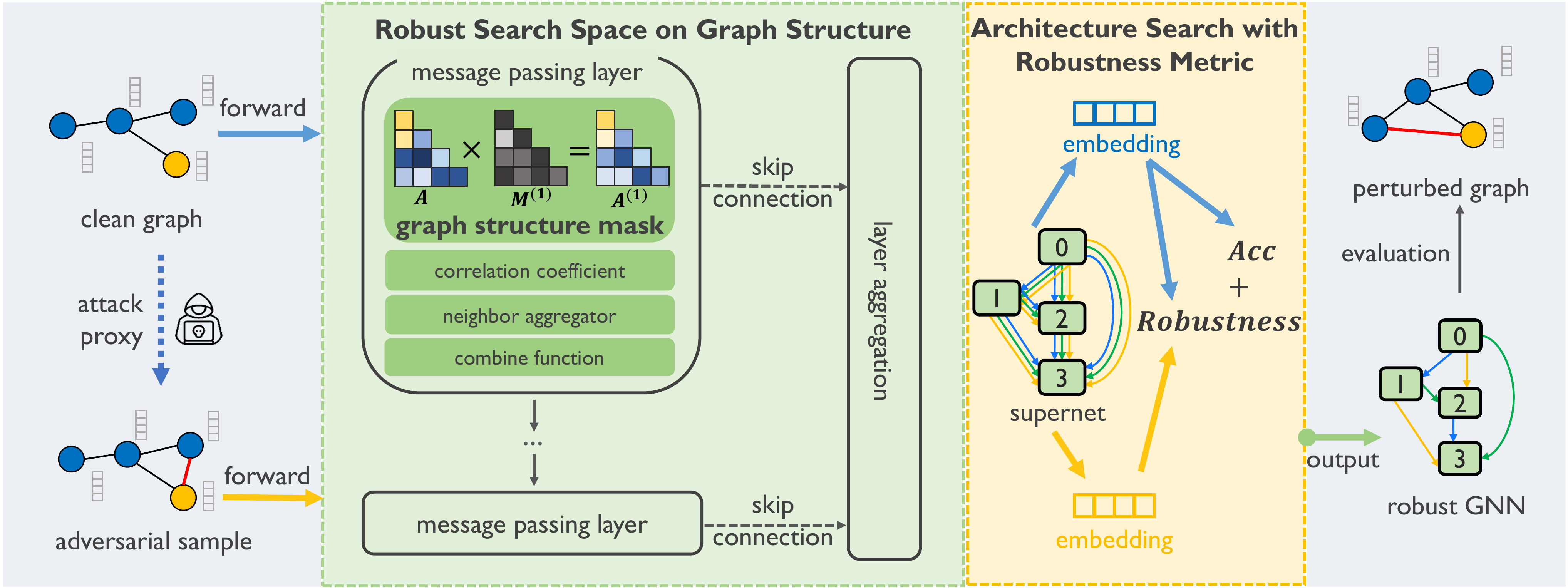}
    \caption{The overall framework of G-RNA. 
    Given a clean graph, the supernet built upon our robust search space is trained in a single-path one-shot way.
    Then, the attack proxy produces several adversarial samples based on the clean graph and we search for robust GNNs with the proposed robustness metric. Finally, we evaluate the optimal robust GNN on graphs perturbed by the attacker.
    }
    \label{fig: framework}
    \vspace{-3mm}
\end{figure*}

In order to reduce human efforts in neural architecture designs, Neural Architecture Search (NAS) has become increasingly popular in both the research field and industry. Though NAS has the potential of automating robust GNN designs, existing graph NAS methods~\cite{ijcai2020-195,2019arXiv190903184Z,DBLP:conf/aaai/LiW0021,Cai_2021_CVPR,SANE} are inevitably susceptible to adversarial attacks since they do not consider adversarial settings and lack robustness designs~\cite{ijcai2021-0637}. Therefore, how to adopt graph NAS to search for optimal robust GNNs in various environments, and in the meantime, fill the gap of understanding GNN robustness from an architectural perspective, remains a huge challenge. 

To address the aforementioned problems and to understand GNN robustness from an architectural perspective, we propose a novel \textbf{R}obust \textbf{N}eural \textbf{A}rchitecture search framework for \textbf{G}raph neural networks (G-RNA), which is the first attempt to exploit powerful graph neural architecture search in robust GNNs, to our best knowledge. 
Specifically, we first design a novel, expressive, and robust search space with graph structure mask operations. 
The green part in \cref{fig: framework} shows the fine-grained search space. 
The graph structure mask operations cover important robust essences of graph structure and could recover various existing defense methods as well.
We train the supernet built upon our designed search space in a single-path one-shot way~\cite{guo2020single}. 
Second, we propose a robustness metric that could properly measure the architecture's robustness. 
Based on the clean graph, an attack proxy produces several adversarial samples.  
We search robust GNNs using our robustness metric with clean and generated adversarial samples. A simple illustration of the robustness metric is shown in the yellow part in \cref{fig: framework}.
After searching for the optimal robust GNN architecture with the evolutionary algorithm, we retrain the top-selected robust architectures from scratch and perform evaluations.
Our contributions are summarized as follows:
\begin{itemize}[leftmargin = 0.3cm]
    \item We develop a robust neural architecture search framework for GNNs, which considers robust designs in graph neural architecture search for the first time to the best of our knowledge. 
    Based on this framework, we can understand adversarial robustness for GNNs from an architectural perspective. 
    \item We propose a novel robust search space by designing defensive operation candidates to automatically select the most appropriate defensive strategies when confronting perturbed graphs.
    Besides, we design a robustness metric and adopt an evolutionary algorithm  together with a single-path one-shot graph NAS framework to search for the optimal robust architectures. 
    \item Extensive experimental results demonstrate the efficacy of our proposed method. G-RNA outperforms state-of-the-art robust GNNs by 12.1\% to 23.4\% on benchmark datasets under heavy poisoning attacks.
\end{itemize}

%% file: 02_related.tex
\section{Related Work}
\label{sec:related}
\subsection{Adversarial Robustness on Graph Data}
Despite the wide success of GNNs in various applications~\cite{zhang2018deep,yugraph,yu2022improving}, GNNs are shown vulnerable to adversarial attacks~\cite{sun2018adversarial,zugner2018adversarial,chang2020restricted,chang2021not,chang2022adversarial,xie2022revisiting}, i.e., slight perturbations to graph can lead to sharp performance decrease.
Following the literature, adversarial attacks have various taxonomies according to the attacker's knowledge (white-box attack and black-box attack), perturbation type (structure attack and feature attack), attack stage (evasion attack and poisoning attack), and targeted nodes (targeted attack and non-targeted attack). 

In response to adversarial attacks, several defensive models have been proposed to enhance the robustness of GNNs. 
Graph pre-processing methods identify and rectify potential structural perturbations before the GNN model training. For example, with the assumption of feature smoothness, GCN-Jaccard~\cite{wu2019adversarial} removes edges that have a low Jaccard similarity. Observing the high-rank tendency of the adjacency matrix under Nettack, GCN-SVD~\cite{entezari2020all} reconstructs the adjacency matrix via its low-rank approximation. 
Graph attention methods aim to learn fewer attention weights on susceptible edges/features. For example,
RGCN~\cite{RGCN} uses Gaussian distribution for hidden layer node representations and calculates attention based on their variance.
Pro-GNN~\cite{jin2020graph} jointly learns graph structure and model parameters by keeping a low-rank and sparse adjacency matrix as well as feature smoothness.
GNN-Guard~\cite{zhang2020gnnguard} learns to focus more on edges between similar nodes and pruning edges between unrelated nodes.
VPN~\cite{jin2021power} defenses by re-weighting edges from graph powering considering $r$-hop neighbors.
Besides, graph certificate robustness methods~\cite{jin2020certified,wang2021certified} provide a theoretical guarantee for graphs to be certified as robust under perturbation budgets. However, this branch is out of the scope of this work and we leave the comparison as a future work.

All previous studies rely heavily on manual designs and thus cannot adapt to new data, tasks, or adversarial attacks. Besides, existing methods neglect the inherent robustness of GNN architectures. 
On the contrary, our proposed method with specifically designed search space on graph structure can automatically search for the optimal robust GNN for different data and tasks.

\subsection{Graph Neural Architecture Search}

Neural Architecture Search (NAS) is a proliferate research direction that automatically searches for high-performance neural architectures and reduces the human efforts of manually-designed architectures. 
NAS on graph data is challenging because of the non-Euclidean graph property and special neural architectures~\cite{you2020design,ijcai2021-0637}.
GraphNAS~\cite{ijcai2020-195} uses the recurrent network as the controller to generate GNN architectures and adopts reinforcement learning to search for optimal architectures.
In order to conduct an efficient search, differentiable NAS approaches~\cite{liu2018darts,guo2020single} jointly optimize the model weights and architecture parameters. DSS~\cite{DBLP:conf/aaai/LiW0021} proposes a differentiable one-shot graph NAS and dynamically updates the search space. 
SANE~\cite{SANE} also utilizes a differentiable search algorithm and builds GNN architectures with the Jumping Knowledge Network (JK-Net) backbone~\cite{Xu2018Representation}. 
GNAS~\cite{Cai_2021_CVPR} reconstructs GNNs with the designed GNN Paradigm and learns the optimal message-passing depth as well.
GASSO~\cite{qin2021graph} uses graph structure learning as a denoising process in the differentiable architecture searching process.
Graph NAS is also used in complex graph data such as heterogeneous graphs~\cite{10.1145/3447548.3467447} and temporal graphs \cite{10.1145/3442381.3449816}. However, the existing graph NAS methods do not consider robustness against adversarial attacks.

\subsection{Robust Neural Architecture Search}
Robust neural architecture search exploits NAS to search for adversarially robust neural architectures.
Since there is no related work for robust NAS on graph data, we review two remotely related papers on computer vision.
RobNets~\cite{guo2020meets} is the first work to explore architecture robustness. Through one-shot NAS, RobNets finetune architecture candidates via adversarial training and then sample more robust architectures. 
DSRNA~\cite{hosseini2021dsrna} proposes two differentiable metrics which help to search robust architectures by a differentiable search algorithm. 
Our work is neither based on adversarial training nor adopts continuous relaxation for architecture parameters.
To conclude, our work differs in that we consider a disparate search space tailored for graph data and leverage a different search algorithm.

%% file: 03_preliminary.tex
\section{Preliminaries}
\label{sec:preliminaries}

\subsection{Graph Neural Networks}
Let $\mathcal{G}=(\mathbf{A},\mathbf{X})$ denote a graph with $N$ nodes, where $\mathbf{A} \in \mathbb{R}^{N\times N}$ is the adjacency matrix and $\mathbf{X}\in \mathbb{R}^{N\times D_0}$ is the corresponding feature matrix. For node $i$, its neighborhood is denoted as $\mathcal{N}(i)$.

Graph Neural Networks take the graph data as input and output node/graph representations to perform downstream tasks like node classification and graph classification. Typically, for node classification tasks with $C$ labels, we calculate:
\begin{gather}
\label{eq:mpnn}
    \mathbf{z}_i =(f_{\bm{\alpha}} \left(
    \mathbf{A},\mathbf{X})
    \right)_i,
\end{gather}
where $\mathbf{z}_i \in \mathbb{R}^{C}$ is the prediction vector for node $i$, $f_{\bm{\alpha}}$ denotes the graph neural network based on architecture $\bm{\alpha}$.
The design of GNNs could be divided into intra-layer design, inter-layer design, and learning configurations~\cite{you2020design}. Intra-layer design often follows the message-passing paradigm~\cite{Gilmer2017Neural}: nodes representation is updated by aggregating neighborhood information. A general formula for updating node representation in GNNs is denoted as
\begin{small}
\begin{gather}
    \mathbf{h}_i^{(l)} =\sigma \left(
    \mathbf{W}^{(l)} \operatorname{Comb} \left(\mathbf{h}_i^{(l-1)}, \operatorname{Aggr}(e_{ij}^{(l)}\mathbf{h}_j^{(l-1)}, j\in \tilde{\mathcal{N}}(i))\right)
    \right),
\end{gather}
\end{small}
where $\mathbf{h}_i^{(l)}$ denotes the node representation for node $i$ in the $l$-th hidden layer, $e_{ij}^{(l)}$ is the correlation coefficient between node $i$ and $j$, $\tilde{\mathcal{N}}(i)=\{i\} \cup \mathcal{N}(i)$ represents the neighborhood of node $i$ with the self-loop, $\operatorname{Aggr}(\cdot)$ is the function to aggregate neighborhood information, $\operatorname{Comb}(\cdot)$ aims for combining self- and neighbor-information, and $\sigma(\cdot)$ is the activation function.

Besides the GNN layer design, how to connect different hidden layers is also critical. Some GNNs directly use the last hidden layer output as the prediction, while pre-processing layers, post-processing layers, and skip connections could also be added~\cite{li2019deepgcns}.
As for learning configurations, they are hyper-parameters for training GNNs like the learning rate, the optimizer, etc.

\subsection{Graph Neural Architecture Search}
In general, graph NAS could be formulated as the following bi-level optimization problem:
\begin{gather}
    \bm{\alpha}^* = \mathop{\arg\max}\limits_{\bm{\alpha} \in \mathcal{O}} \operatorname{Acc}_{val}(\bm W^*(\bm{\alpha}), \bm{\alpha}) 
    \label{eq: up}\\
    \operatorname{ s.t. } \ \bm W^*(\bm{\alpha}) = \mathop{\arg\min}\limits_{\bm W}\mathcal{L}_{train}(\bm W,\bm{\alpha})\label{eq: low},
\end{gather}
where Eq.~\eqref{eq: up} is the upper-level optimization problem to find the best architecture $\bm{\alpha}^*$ in the search space $\mathcal{O}$, and Eq.~\eqref{eq: low} is the lower-level problem to calculate optimal model weights $\bm W$ for one particular architecture $\bm{\alpha}$. 
$\operatorname{Acc}_{val}$ represents the prediction accuracy on the validation set and $\mathcal{L}_{train}$ is the classification cross-entropy loss on the training set.

Existing graph NAS methods~\cite{ijcai2021-0637,guan2021autogl,2019arXiv190903184Z,ijcai2020-195,Cai_2021_CVPR} design their search space following the message-passing scheme in Eq.~\eqref{eq:mpnn}.
The most commonly used correlation coefficient operations are provided in \cref{appendix: cor}. NAS methods can search all components in Eq.~\eqref{eq:mpnn} such as the aggregation function, correlation coefficients, and activation functions as well as hyper-parameters like hidden size and learning configurations. In this paper, we mainly consider searching for architectural designs.

%% file: 04_method.tex
\section{Robust Graph Neural Architecture Search}
\label{sec:method}
In this section, we first formulate the problem of graph robust NAS. 
Then, we introduce our novel and expressive search space with defensive operation candidates, namely graph structure masks.
Based on the designed search space, we build a supernet containing all possible architectures and train it in a single-path one-shot way. 
Finally, we introduce our proposed robustness metric and describe the search process exploiting the evolutionary algorithm.

\subsection{Problem formulation}
Given a search space $\mathcal{O}$, we aim to find the optimal architecture $\bm{\alpha}^* \in \mathcal{O}$ with both high prediction accuracy and high adversarial robustness. 
We formulate the problem of robust neural architecture search for GNNs as:
\begin{align}
    \bm{\alpha}^* = \mathop{\arg\max}\limits_{\bm{\alpha} \in \mathcal{O}} \operatorname{ACC}_{val}(\bm{\alpha}) + \lambda \mathcal{R}(\bm{\alpha}), 
    \vspace{-5mm}
\end{align}
where $\mathcal{R}(\cdot)$ is the robustness metric, and $\lambda$ is a hyper-parameter balancing the model accuracy and robustness.

\subsection{Search Space for Robust GNNs}
We design a fine-grained  search space following the message-passing paradigm. In total, there are six adjustable components in our GNN architecture: the graph structure mask, the nodes correlation coefficient, the neighbor aggregator, the combine function, the skip connection, and the layer aggregator. The first four components belong to the intra-layer operations, while the rest two components are inter-layer operations.

\xhdr{Intra-layer Operations}
Inside the $l$-th message-passing layer, the defensive operation $\mathcal{D} \in \mathcal{O}_\mathcal{D}$ is firstly adopted to construct a graph structure mask and reconstruct the graph structure: 
\begin{equation}
    \mathbf{M}^{(l)} = \mathcal{D}(\mathbf{A}^{(l-1)}), \\
    \mathbf{A}^{(l)} = \mathbf{A} \odot \mathbf{M}^{(l)}
\end{equation}
where $\mathbf{M}^{(l)}=\{m_{ij}\}$ is the graph structure mask matrix and $\mathbf{A}^{(l)}$ denotes the graph structure in the $l$-th layer. 
$\odot$ is the Hadamard product and $\mathbf{A}^{(0)} = \mathbf{A}$. 
Each element $m_{ij} \in [0,1]$ is the mask score between node $i$ and node $j$ where $m_{ij}=0$ indicates a complete edge pruning and $m_{ij}=1$ means no modification to the original edge. 
The defensive operation aims to assign fewer weights to potential perturbed edges.

\begin{table}[t]
    \centering
    \caption{Graph structure mask operations $O_{\mathcal{D}}$. \revision{The detailed denotation is introduced in~\cref{app:denotation}.}}
\label{table: robust operations}
    \begin{tabular}{l|l}
    \toprule
    $\mathcal{O}_{\mathcal{D}}$ &  \textbf{Formula} \\
    \midrule
    \textit{Identity} & $\mathbf{M}^{(l)}=\mathbf{A}^{(l-1)}$\\
    \midrule
    \multirow{2}{*}{\textit{LRA}} & 
    $\mathbf{A}^{(l-1)}=\mathbf{U}^{(l-1)} \mathbf{S}^{(l-1)} (\mathbf{V}^{(l-1)})^T,$ \\ 
    & $\mathbf{M}^{(l)}=\mathbf{U}_r^{(l-1)} \mathbf{S}_r^{(l-1)} (\mathbf{V}_r^{(l-1)})^T$
    \\ 
    \midrule
    \textit{NFS}    &  $m_{ij}^{(l)} = \left\{
    \begin{aligned}
         0,\quad & if \ a_{ij}^{(l-1)} > 0  \ and \ J_{ij}< \tau  \\
         a_{ij}^{(l-1)}, \quad &otherwise
    \end{aligned}
    \right.$  \\
    \midrule
    \textit{NIE}   &  $m_{ij}^{(l)} = \beta m_{ij}^{(l-1)} + (1-\beta) \hat{\alpha}_{ij}^{(l)} $ \\
    \midrule
    \textit{VPO}  &  $\mathbf{M}^{(l)} = \sum_{v=1}^V \theta_v (\mathbf{A}^{(l-1)})^v$\\
    \bottomrule
    \end{tabular}
    \vspace{-4mm}
\end{table}

Inspired by the success of current defensive approaches~\cite{sun2018adversarial}, we conclude the properties of operations on graph structure for robustness and 
design representative defensive operations in our search space accordingly.
In this way, we can choose the most appropriate defensive strategies when confronting perturbed graphs. To our best knowledge, this is the first time the search space to be designed with a specific purpose to enhance the robustness of GNNs.
Specifically, we include five graph structure mask operations in the search space.
\textit{Identity} keeps the same graph structure as the previous layer.
\textit{L}ow \textit{R}ank \textit{A}pproximation (\textit{LRA}) reconstructs the adjacency matrix in the $l$-th layer from the top-$r$ components of singular value decomposition from adjacency matrix in the previous layer. 
\textit{N}ode \textit{F}eature \textit{S}imilarity (\textit{NFS}) deletes edges that have small Jaccard similarities among node features. 
\textit{N}eighbor \textit{I}mportance \textit{E}stimation (\textit{NIE}) updates mask values with a pruning strategy based on quantifying the relevance among nodes. 
\textit{V}ariable \textit{P}ower \textit{O}perator (\textit{VPO})
    forms a variable power graph from the original adjacency matrix weighted by the parameters of influence strengths.

Moreover, more operations like graph structure learning~\cite{GSL_survey} could be integrated into the graph structure mask operations.
The formula for mask operations is shown in~\cref{table: robust operations}.
We limit \textit{LRA} and \textit{NFS} in the first layer (only for pre-processing) and exploit the other three operations for all layers.
It is worth mentioning that the graph structure mask candidates could be easily extended to other methods that deal with the graph structure, like denoising methods.
Mask operations will not occlude the later message-passing process. They could be seen as orthogonal operations to other operations like the correlation coefficient. 

\begin{table}[htbp]
    \centering
    \caption{The six components in search space and corresponding candidate operations.}
    \vspace{-2mm}
    \label{table: search space}
    \begin{tabular}{l|l}
    \toprule
    \textbf{Component} &  \multicolumn{1}{l}{\textbf{Candidate Operations}} \\
    \midrule
    $\mathcal{O_D}$ & \begin{tabular}[c]{@{}l@{}}
    \textit{Identity, LRA, NFS, VPO}, \textit{NIE}\end{tabular}                       \\
    \midrule
    $\mathcal{O}_e$ & \begin{tabular}[c]{@{}l@{}}\textit{Identity, GCN, GAT, GAT-Sym, Cos}, \\ \textit{Linear, Gene-Linear}\end{tabular} \\
    \midrule
    $\mathcal{O}_{aggr}$     & \textit{Sum, Mean, Max}                                                    \\
    \midrule
    $\mathcal{O}_{comb} $    & \textit{Identity, GIN, SAGE} \\
    \midrule
    $\mathcal{O}_{skip}$ & \textit{Identity, Zero}  \\
    \midrule
    $\mathcal{O}_{layer}$    & \textit{Concat, Max, LSTM} \\
    \bottomrule
    \end{tabular}
\vspace{-2mm}
\end{table}

For the other three components in the intra-layer, our choices for correlation coefficients follow the literature summarized in~\cref{table: correlation coefficient} in Appendix.
Based on the masked graph structure $\mathbf{A}^{(l)}$ and correlation coefficients $\{e_{ij}^{(l)}\}$, a neighbor aggregator $\operatorname{Aggr} \in \mathcal{O}_{aggr}$ is used to aggregate neighborhood representations. 
Afterwards, a combine function $\operatorname{Comb} \in \mathcal{O}_{comb}$ merges self- and neighbor-message. 
Here, we explore two typical approaches to combine self-representation and neighbor-message, namely \textit{GraphSAGE}~\cite{Hamilton2017Inductive} and \textit{GIN}~\cite{xu2019how}.
\textit{GraphSAGE} conducts different feature transformations for node representation and neighborhood information. \textit{GIN} first performs weighted sum for self- and neighbor-message, then uses multi-layer perceptron to improve GNN's expressive power. 
After combining messages, the activation function $\sigma(\cdot)$ is applied. 
Overall, the hidden representation for node $i$ in the $l$-th layer is calculated as 

\begin{footnotesize}
\begin{equation}
    \mathbf{h}_i^{(l)} =\sigma \left(
    \mathbf{W}^{(l)} \operatorname{Comb} \left(\mathbf{h}_i^{(l-1)}, \operatorname{Aggr}(m_{ij}^{(l)}e_{ij}^{(l)}\mathbf{h}_j^{(l-1)}, j\in \tilde{\mathcal{N}}(i))\right)
    \right).
\end{equation}
\end{footnotesize}
The node representation is initialized as node features, i.e., $\mathbf{h}^{(0)}=\mathbf{X}$.

\xhdr{Inter-layer Operations}
Following the idea of JK-Net~\cite{Xu2018Representation} and SANE~\cite{SANE}, we aggregate node representations in intermediate layers via the layer aggregator $\operatorname{Layer\_Aggr} \in \mathcal{O}_{layer}$. The skip operation $\operatorname{Skip} \in \mathcal{O}_{skip}$ decides the skip connection to the final layer aggregator. 
\begin{align}
    \mathbf{z}_{i}=\text { Layer\_Aggr }\left(
    \text{Skip}(\mathbf{h}_{i}^{(1)}),...,\text{Skip}( \mathbf{h}_{i}^{(L)})
    \right),
\end{align}
where $L$ is the maximum number of message-passing layers. 
With the JK-Net backbone, we choose the optimal model depth by setting a maximal depth.
For example, if we set $L=4$ and the skip operations are [\textit{Idenity, Zero, Idenity, Zero}], then the selected optimal number of message-passing layers is 3.
After obtaining the final node representations, we add several fully connected layers to conduct classification.

We summarize all components and their candidate operations in~\cref{table: search space}. 
Our search space is expressive and, more importantly, has improved defense capability. As shown in~\cref{table: GNNs} in Appendix, our search space could recover some classic manually designed GNNs and state-of-the-art robust GNNs like GCN-SVD, GCN-Jaccard, GNN-Guard, and VPN.

\xhdr{Supenet Training}
With the proposed search space, we construct a supernet and train it in a single-path one-shot way~\cite{guo2020single}.
The supernet we built contains all possible architectures based on our search space, also called the one-shot model. In the supernet, each architecture behaves as a single path.
Unlike graph NAS methods such as GNAS~\cite{Cai_2021_CVPR} or DSS~\cite{DBLP:conf/aaai/LiW0021} that use continuous relaxation for architecture parameters, we train the supernet via uniform single path sampling~\cite{guo2020single}.
During the training process, we uniformly sample one path to update weights each time. Afterward, we search for optimal architectures through the evolutionary algorithm without further training steps.

\subsection{Measuring Robustness}
This part will introduce how to measure the architecture's robustness and the specific search process using the evolutionary algorithm with our robustness metric.

Intuitively, the performance of a robust GNN should not deteriorate too much when confronting various perturbed graph data. 
Let $\mathcal{T}_\Delta$ denote the attacker of the graph data with perturbation budgets $\Delta$. We define the robustness metric as
\begin{small}
\begin{equation}
\label{eq:robustmetric}
     \mathcal{R}(\mathbf{A},f) = -\mathbb{E}_{\mathbf{A}'} \left[\frac{1}{N}\sum_{i=1}^N D_{KL}\left( f(\mathbf{A})_i||f(\mathbf{A}')_i\right) \right], 
     \mathbf{A}' = \mathcal{T}_{\Delta}(\mathbf{A}),
\end{equation}
\end{small}
where $f$ indicates the GNN, $f(\mathbf{A})_i$ is the probability prediction vector for node $i$ and $D_{KL}(p||q) = \sum_i p_i \log \frac{p_i}{q_i}$ denotes the Kullback-Leibler (KL) divergence between two distribution $p$ and $q$. 
Here, we use KL distance to measure the prediction difference given clean and perturbed data.
The choice of our attack proxy $\mathcal{T}$ varies from simple random attack~\cite{dai2018adversarial} or DICE~\cite{zugner_adversarial_2019} to advanced attack methods like Mettack~\cite{zugner_adversarial_2019} or Nettack~\cite{zugner2018adversarial}. 
Within the same perturbation budgets $\Delta$, our attack proxy generates $T$ perturbed graph structure $\{\mathbf{A}'_t\}_{t=1}^T$. 
Eq.~\eqref{eq:robustmetric} could be approximately computed as
\begin{align}
    \mathcal{R}(\mathbf{A},f) \approx -\frac{1}{TN}\sum_{t=1}^{T}\sum_{i=1}^{N} \left(D_{KL}( f(\mathbf{A})_i||f(\mathbf{A}_t')_i) \right).
\end{align}
A larger $\mathcal{R}(\mathbf{A},f)$ indicates a smaller distance between the prediction of clean data and the perturbed data, and consequently, more robust GNN architectures.

\begin{figure}[htbp]
    \centering 
    \begin{subfigure}[b]{0.47\linewidth}
     \centering 
      \includegraphics[width=1\textwidth]{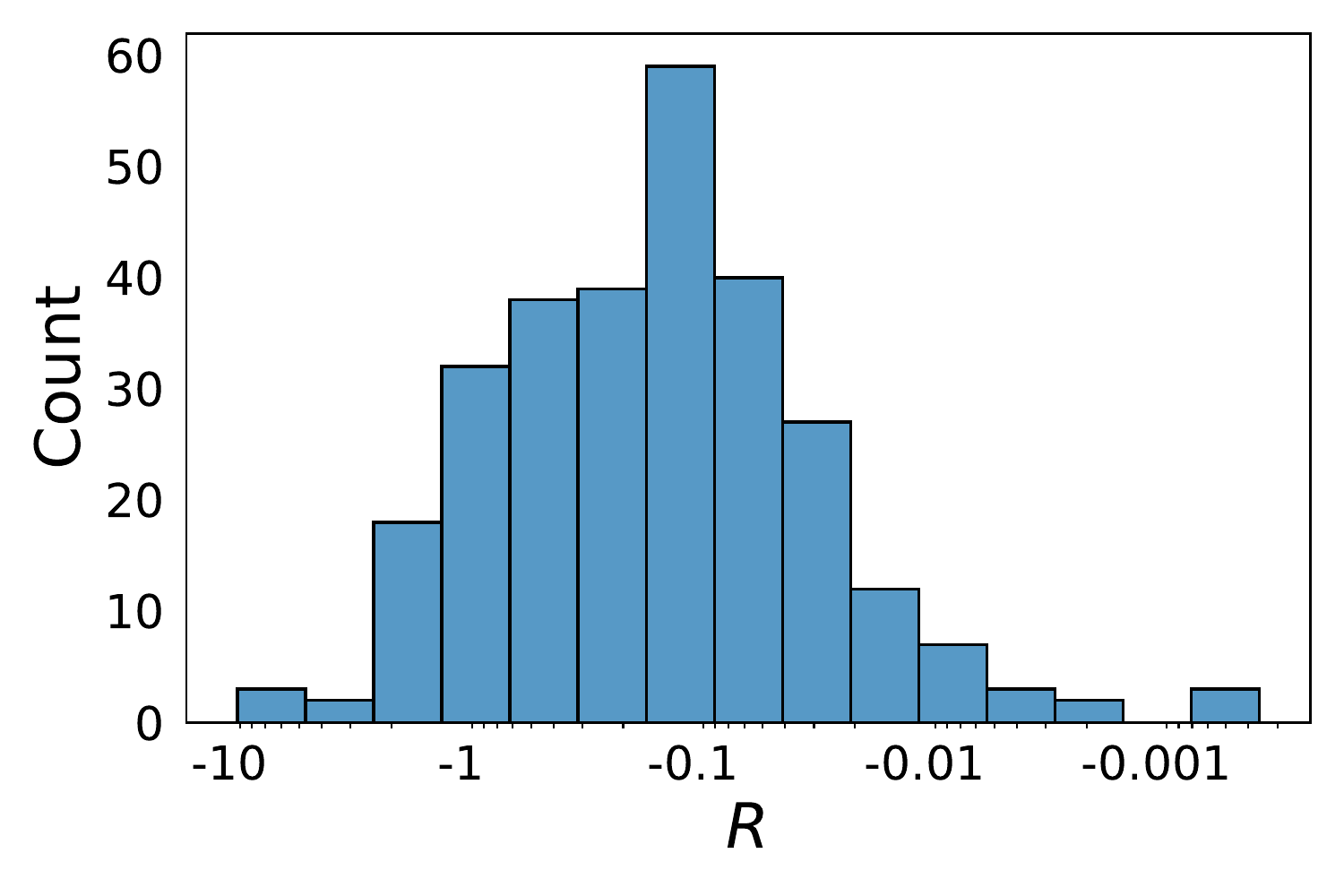}
    \caption{}\label{fig:hist}
    \end{subfigure}
    \begin{subfigure}[b]{0.47\linewidth}
     \centering 
    \includegraphics[width=1\textwidth]{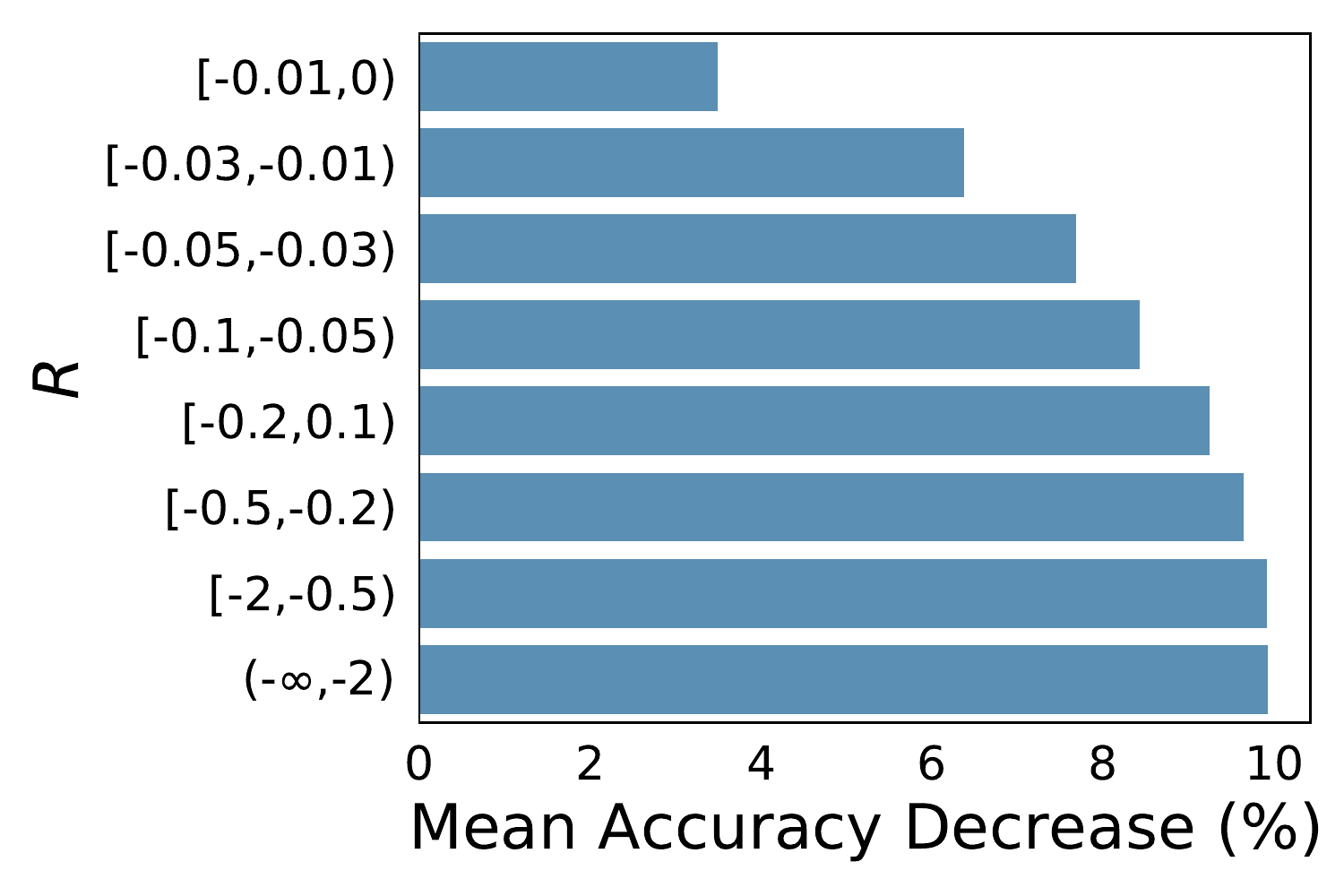}
    \caption{}
    \label{fig:rob}
    \end{subfigure}
    \vspace{-2mm}
    \caption{Evaluation of the robustness metric $\mathcal{R}$. (a) A histogram for the robustness metric in log scale. (b) The relationship between the robustness metric and accuracy decreases (\%) under attacks.} 
    \vspace{-5mm}
 \end{figure}
 
In order to illustrate the effectiveness of the proposed robustness metric, we randomly choose 300 architectures and calculate their robustness metric value and evaluate their performance decrease after 5\% structural perturbations in the Cora dataset (the experimental details are described in \cref{exp: rob}). \cref{fig:hist} displays the distribution of $\mathcal{R}$ in log scale while \cref{fig:rob} shows the mean accuracy decrease for different $\mathcal{R}$ intervals. The accuracy decrease could be regarded as a ground-truth measurement of the actual robustness.
From \cref{fig:rob}, we could see that the mean accuracy decrease has a negative correlation with respect to the robustness metric.
When the robustness metric is relatively small, the accuracy may be reduced by as large as 10\%. 
However, when the robustness metric is large, the accuracy decrease shrinks to about 3.5\%. This phenomenon indicates that our robustness metric could successfully filter robust architectures.

\xhdr{Evolutionary Search Algorithm}
In our work, we adopt an evolutionary algorithm to search for optimal robust graph architectures with the proposed robustness metric.
Instead of training each candidate architecture from scratch, we leverage the evolutionary algorithm only for inference and search. The weights for all architectures are fixed as those learned in the supernet training phase. 
In one search epoch, we select top-$k$ robust candidates via the fitness function $\operatorname{ACC}_{val}(\bm{\alpha}) + \lambda \mathcal{R}(\bm{\alpha})$. Crossover and mutation are followed to generate children architectures from population candidates. 
We show the search algorithm in~\cref{appendix: ea}.

%% file: 05_experiments.tex
\section{Experiments}
\label{sec:experiments}
In this section, we conduct experiments to verify our proposed method by evaluating the selected architecture on perturbed graphs.
Also, we visualize the performance for diverse operations to better understand GNN architecture's robustness. 
Additional experimental results including defensive performance under targeted attack, evaluation on heterophily graphs, and sensitivity analysis for the hyper-parameter $\lambda$ in our proposed robustness metric could be found in~\cref{addition_exp}.
The details of the experimental setting are deferred to the \cref{appendix: Experiment Details}. 

\begin{table*}[t]
\caption{The results of node classification accuracy (mean$\pm$std, in percentages) under non-targeted attacks (Mettack).
\textbf{Bold} numbers indicate the best performance. 
``-'' indicates the result is unavailable due to the high time complexity of the model.}
\label{tab:result}
\centering
\resizebox{1.95\columnwidth}{!}{
\begin{tabular}{ccccccccc}
\toprule
\multirow{2}{*}{Dataset} & \multicolumn{2}{c}{\multirow{2}{*}{Model}} & \multicolumn{6}{c}{Proportion of changed edges (\%)} \\ \cmidrule(r){4-9} 
 & \multicolumn{2}{c}{} & 0 & 5 & 10 & 15 & 20 & 25 \\
 \midrule
\multirow{13}{*}{Cora} & \multirow{4}{*}{Vanilla GNN} & GCN & 84.28±0.25 & 78.00±1.20 & 70.31±1.24 & 56.97±1.20 & 48.56±2.66 & 43.83±1.47 \\
 &  & GCN-JK & 84.38±0.38 & 74.63±0.60 & 67.32±0.80 & 53.26±1.17 & 45.29±2.49 & 38.63±1.03 \\
 &  & GAT & 84.30±0.65 & 78.56±1.33 & 70.18±1.43 & 58.39±2.27 & 49.35±1.55 & 42.40±1.02 \\
 &  & GAT-JK & 84.16±0.36 & 74.30±1.35 & 68.07±1.12 & 54.54±2.55 & 51.20±1.51 & 43.29±1.47 \\ 
\cmidrule(r){2-9} 
 & \multirow{5}{*}{Robust GNN} & RGCN & 84.60±0.37 & 75.92±1.01 & 72.94±0.40 & 59.97±0.50 & 52.50±0.38 & 46.47±0.91 \\
 &  & GCN-Jaccard & 83.64±0.76 & 77.07±0.61 & 74.07±0.59 & 68.92±0.80 & 63.57±0.87 & 56.14±1.45 \\
 &  & Pro-GNN & 84.64±0.59 & 79.59±0.83 & 73.73±0.76 & 62.10±1.65 & 54.89±2.03 & 48.98±1.89 \\
&  &DropEdge & 83.35±1.23 & 77.84±0.30 & 70.96±0.44 & 57.13±0.50 & 49.70±0.72 & 43.51±1.13 \\
  &  &PTDNet & 83.70±0.43 & 78.49±0.43 & 70.94±0.34 & 54.39±0.81 & 45.80±0.63 & 41.32±0.67 \\
 \cmidrule(r){2-9} 
  & \multirow{4}{*}{Graph NAS} & GraphNAS & 82.77±0.40 & 72.97±2.34&	57.12±5.31&	44.50±1.48	&37.21±3.79	& 31.96±1.68 \\
   &  & GASSO & 84.11±0.34 & 77.69±1.10 & 68.50±0.64 & 56.61±0.90 & 51.87±0.79 & 46.05±2.01 \\
 &  & {G-RNA w/o rob} & \textbf{84.29±0.40} & 77.39±1.38 & 67.61±1.73 & 53.56±3.00 & 48.57±1.85 & 41.20±1.53 \\
 & & \textbf{G-RNA} & 83.81±0.39 & \textbf{80.45±0.74} & \textbf{75.16±0.89} & \textbf{73.52±0.86} & \textbf{70.6±1.43} & \textbf{67.23±1.66} \\
 \midrule
\multirow{13}{*}{CiteSeer} & \multirow{4}{*}{Vanilla GNN} & GCN & 72.35±0.49 & 63.48±0.45 & 56.94±1.29 & 55.01±0.91 & 43.73±1.15 & 40.47±0.77 \\
 &  & GCN-JK & 72.14±0.36 & 62.16±0.51 & 54.25±1.30 & 50.61±0.93 & 40.06±1.11 & 35.07±1.51 \\
 &  & GAT & 71.75±0.71 & 61.47±1.33 & 53.92±1.29 & 49.68±2.74 & 41.31±2.95 & 37.25±2.26 \\
 &  & GAT-JK & 71.52±0.90 & 63.90±0.38 & 57.16±0.53 & 52.80±1.08 & 44.26±0.87 & 39.63±0.69 \\ 
 \cmidrule(r){2-9} 
 & \multirow{5}{*}{Robust GNN} & RGCN & 72.43±0.41 & 63.30±0.33 & 55.90±0.47 & 53.83±0.31 & 43.65±0.72 & 39.99±0.46 \\
 &  & GCN-Jaccard & 71.03±0.45 & 64.56±0.75 & 57.57±0.74 & 55.31±0.82 & 50.17±0.66 & 45.78±0.43 \\
 &  & Pro-GNN & 71.74±0.76 & 66.29±0.64 & 64.60±0.86 & 63.96±1.48 & 62.46±1.12 & 55.73±2.04 \\ 
 &  &{DropEdge} & 71.84±0.21 & 63.33±0.70 & 55.41±0.77 & 50.81±0.93 & 40.61±0.92 & 37.46±0.61 \\
   &  &{PTDNet} & \textbf{72.87±0.46} & 64.08±0.78 & 57.03±0.37 & 54.04±0.75 & 41.81±0.63 & 38.93±0.66  \\
\cmidrule(r){2-9} 
  & \multirow{4}{*}{Graph NAS} & GraphNAS & 72.79±0.22	& 61.01±0.36& 	53.55±1.29& 	55.98±12.01& 	39.06±6.79& 36.90±4.76 \\
  &  & {GASSO} & 71.08±0.29 & 61.31±0.53 & 52.17±0.70 & 50.43±0.87 & 43.72±1.10 & 36.84±0.55 \\
&  & {G-RNA w/o rob} &72.57±0.29 & 65.48±1.29 & 56.59±1.76 & 55.81±1.59 & 48.53±2.20 & 43.76±2.23 \\
 &  & \textbf{G-RNA} & 
71.32±0.82 & \textbf{68.71±1.20} & \textbf{65.84±1.20} & \textbf{65.29±1.35} & \textbf{62.58±0.99} & \textbf{61.33±1.35} \\
 
 \midrule
\multirow{13}{*}{PubMed} & \multirow{4}{*}{Vanilla GNN} & GCN & 86.35±0.15 & 82.70±0.13 & 80.56±0.16 & 77.85±0.17 & 75.85±0.18 & 73.68±0.22 \\
 &  & GCN-JK & 87.07±0.12 & 82.76±0.15 & 81.56±0.18 & 80.22±0.38 & 79.14±0.44 & 77.31±0.24 \\
 &  & GAT & 85.28±0.20 & 81.02±0.31 & 79.58±0.16 & 76.39±0.43 & 74.41±0.20 & 72.22±0.24 \\
 &  & GAT-JK & 85.72±0.14 & 82.37±0.10 & 80.60±0.23 & 78.50±0.15 & 76.39±0.14 & 74.02±0.25 \\ 
 \cmidrule(r){2-9} 
 & \multirow{5}{*}{Robust GNN} & RGCN & 86.64±0.08 & 82.90±0.18 & 80.73±0.19 & 77.86±0.17 & 75.89±0.15 & 73.74±0.22 \\
 &  & GCN-Jaccard & 87.11±0.04 & 83.95±0.06 & 82.30±0.08 & 80.16±0.07 & 78.83±0.13 & 76.86±0.17 \\
 &  & Pro-GNN & - & - & - & - & - & - \\ 
 &  & {PTDNet} &83.87±0.24 & 74.32±0.44 & 68.80±0.34 & 67.32±0.18 & 66.50±0.12 & 65.21±0.34  \\
 &  & {DropEdge} & 83.93±0.10 & 83.24±0.12 & 82.33±0.15 & 81.06±0.18 & 79.21±0.14 & 76.88±0.28 \\
 \cmidrule(r){2-9} 
 & \multirow{4}{*}{Graph NAS} & GraphNAS & 87.26±0.04&	83.56±0.08&	 80.00±3.98&	 77.86±2.59	& 72.97±3.88&	 68.05±2.26 \\
 &  & {GASSO} & 86.27±0.12 & 84.15±0.15 & 83.18±0.21 & 82.56±0.25 & 81.73±0.36 & 83.25±1.26 \\
 &  & {G-RNA w/o rob} & 87.18±0.07 & 82.59±0.14 & 80.29±0.17 & 78.11±0.24 & 75.98±0.33 & 73.60±0.25 \\
 &  & \textbf{G-RNA} & \textbf{87.48±0.12} & \textbf{87.01±0.11} & \textbf{86.5±0.14} & \textbf{86.04±0.21} & \textbf{85.94±0.18} & \textbf{85.82±0.12} \\
  \bottomrule
\end{tabular}
}
\end{table*}

\subsection{Semi-supervised Node Classification Task}
In the semi-supervised node classification task, we perform non-targeted poisoning structural attacks adopting Mettack~\cite{zugner_adversarial_2019} and evaluate GNN robustness based on perturbed graphs.
We vary the proportion of changed edges from 0\% to 25\% and calculate the retrained accuracy on perturbed graphs. 
For each setting, we experiment 10 times and report the average results and the standard deviation. The final defense results are summarized in~\cref{tab:result}. 
More information on searched architectures by G-RNA under attacks is provided in~\cref{appendix: arch}.

Overall, our G-RNA successfully outperforms all baselines under adversarial attacks. We could see an apparent increase in perturbation performance when confronting heavily poisoning attacks (\eg~when the proportion of changed edges is 15\% or more).
For instance, the defense performance of G-RNA under 25\% structural perturbations exceeds that of vanilla GCN by \revision{23.4\%, 20.9\%, and 12.1\%}, respectively.
For NAS-based methods, optimal architectures searched on clean data are evaluated on both clean and perturbed graphs to test their robustness and conduct a fair comparison. We could see that there does exist a trade-off between architecture accuracy and robustness.
Though GraphNAS shows impressive results on clean graphs, its vulnerability under adversarial attacks is also obvious, \eg~GraphNAS's performance under adversarial attacks is even worse than that of classic GNNs sometimes. Thus, it's necessary to make Graph NAS more robust. In CiteSeer, the performance for G-RNA under 20\% and 25\% structural perturbations are almost twice that of GraphNAS, which shows the strong defense ability of our method.

\xhdr{Ablation study on robustness metric} We add an ablation study by removing the robustness metric (denoted as G-RNA w/o rob). Our full G-RNA shows better performance compared to G-RNA w/o rob on all datasets under attacks, demonstrating the effectiveness of our proposed robustness metric.

 \begin{figure}[htbp]
    \centering 
    \begin{subfigure}[b]{0.48\linewidth}  
    \centering \includegraphics[width=1\textwidth]{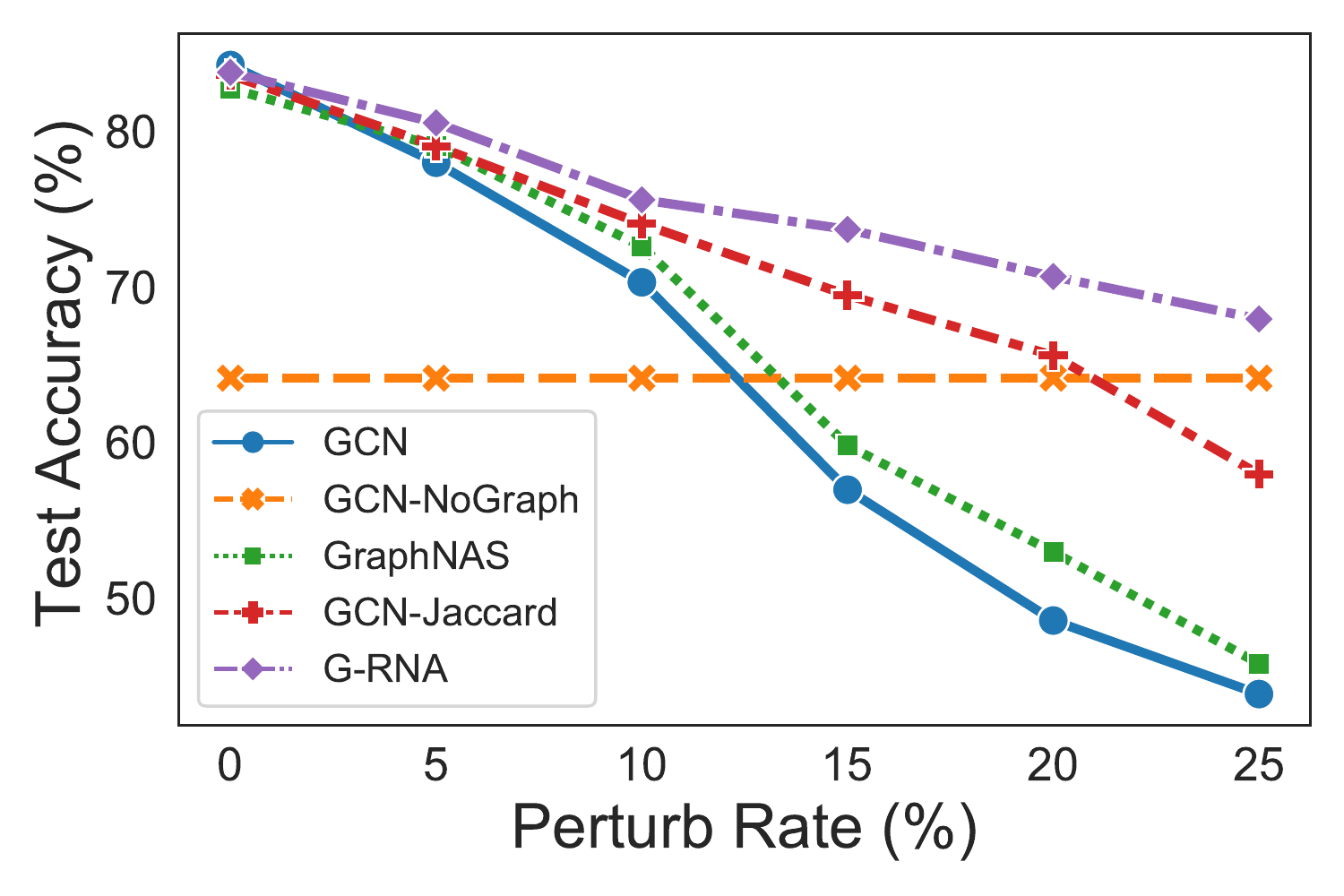} 
    \caption{Cora.}
    \label{fig:nograph subfig cora}
    \end{subfigure}
    \begin{subfigure}[b]{0.48\linewidth} 
    \centering 
    \centering
    \includegraphics[width=1\textwidth]{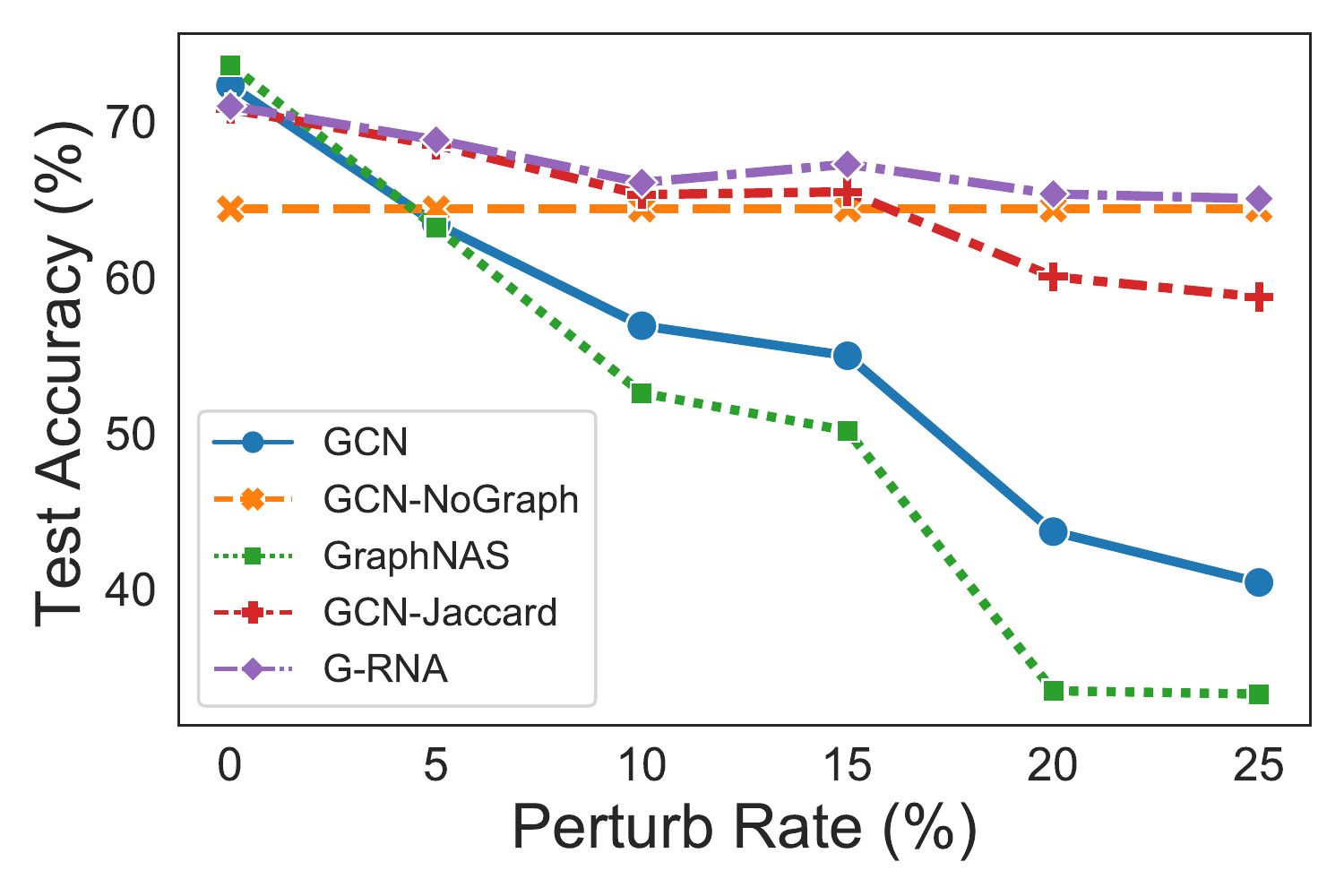}
    \caption{CiteSeer.}\label{fig:nograph subfig citeseer}
    \end{subfigure}
    \vspace{-0.3cm} 
    \caption{Comparison with GCN-NoGraph. G-RNA is the only method that consistently outperforms GCN-NoGraph.} 
    \label{fig: nograph} 
    \vspace{-3mm}
 \end{figure}

\xhdr{Comparison with not using structures}
When the graph structure is heavily poisoned, aggregating neighbor messages by the message-passing paradigm may not be reliable. As a result, we compare our method with GCNs not using edges, named GCN-NoGraph~\cite{jin2020graph}, i.e., a two-layer MLP on node features.
Notice that the performance exceeding GCN-NoGraph indicates an effective graph structure, while that below GCN-NoGraph implies an informationless or confusing graph structure for GNNs. 
\cref{fig: nograph} shows the comparison with GCN-NoGraph. 
For all comparing methods, only the performance of G-RNA always outperforms GCN-NoGraph.
GCN-Jaccard performs well for small perturbations but fails to defend against large perturbations. 
Also, we could see an obvious drop for GCN and GraphNAS when the perturb rate goes beyond 15\% for both datasets.

\subsection{Understand Operation Robustness}\label{exp: rob}
\begin{figure*}[htbp]
    \centering 
    \vspace{-0.6cm}
    \begin{subfigure}[b]{1.\linewidth}
    \centering
    \includegraphics[width=\textwidth]{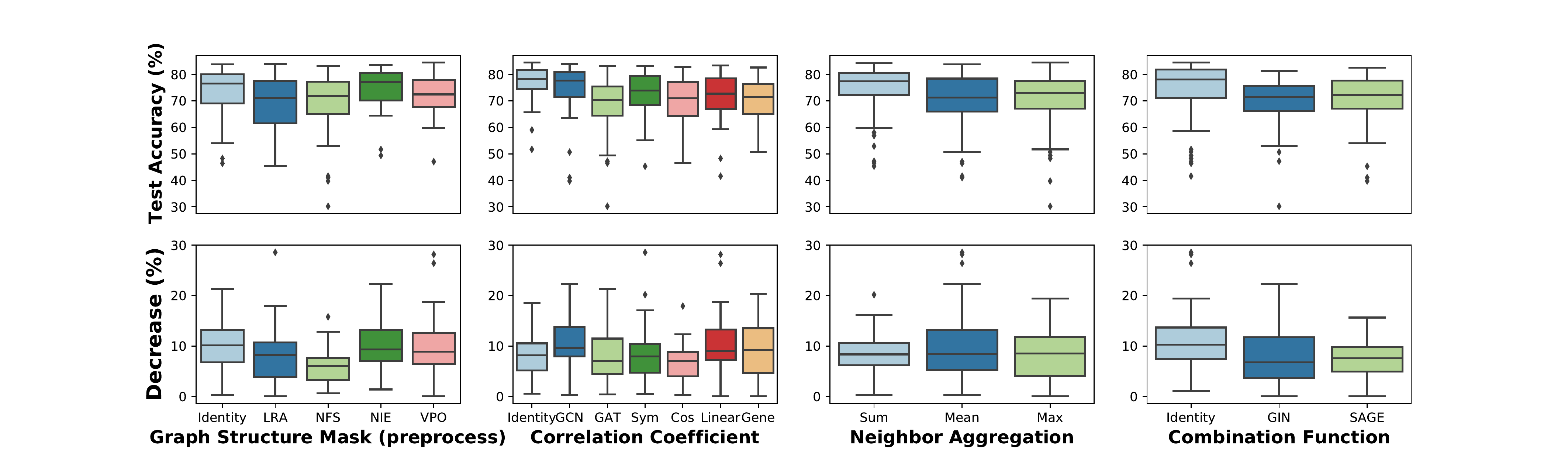}
    \caption{Intra-layer importance analysis.}
    \label{fig: intra}
    \end{subfigure}%
    \\
    \vspace{-2mm}
    \begin{subfigure}[b]{1\linewidth}
    \centering
    \includegraphics[width=0.7\textwidth]{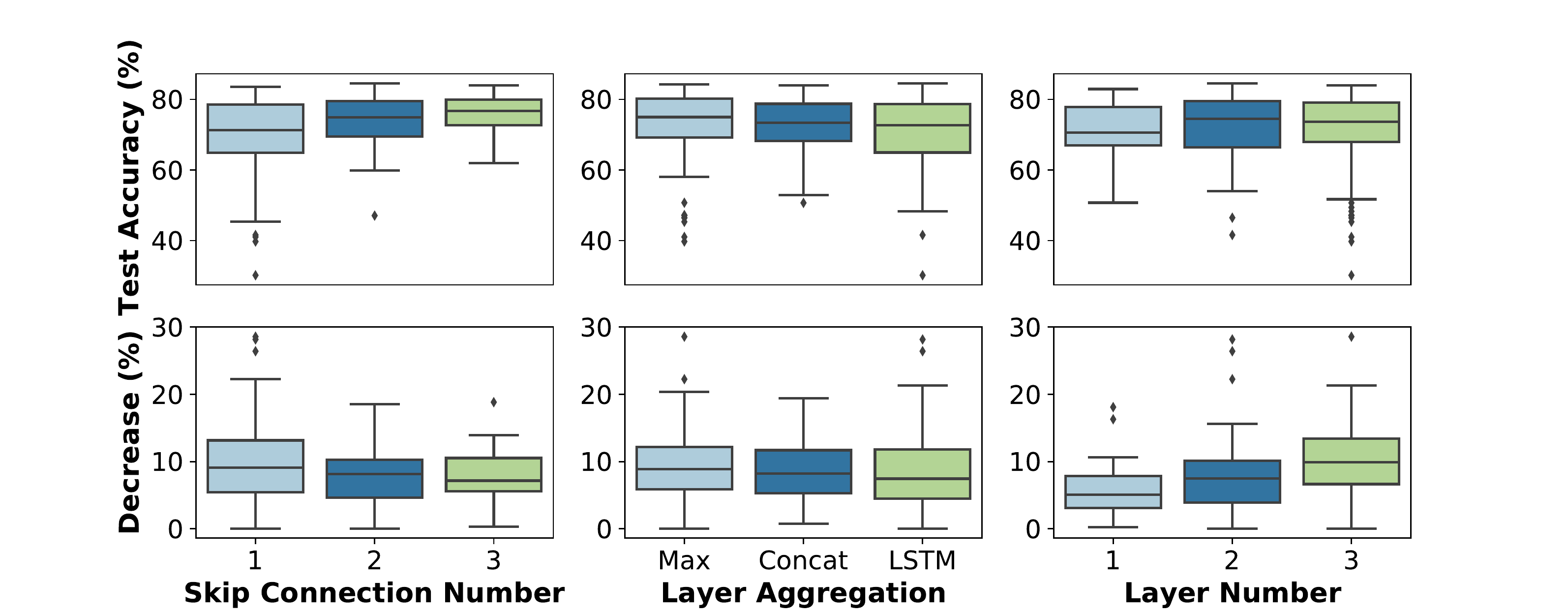}
    \caption{Inter-layer importance analysis.}
    \label{fig: inter}
    \end{subfigure}
    \vspace{-0.5cm} 
    \caption{The importance of intra-layer and inter-layer designs to GNN robustness. The first row of each picture is the clean test accuracy (\%) for various operations(higher is better), while the second row shows the accuracy decrease (\%) under perturbations (lower is better).} 
    \vspace{-4mm}
 \end{figure*}

For understanding the robustness of various GNN components, we randomly sample 300 architectures and evaluate their performance on clean and attacked graphs.
To make a fair comparison, we keep the same training configurations for all selected architectures. Specifically, we train each architecture for 200 epochs with a learning rate of 0.005 and a weight decay of 5e-4.
We use the decrease in classification accuracy as the measurement and report the results on the Cora dataset under Mettack (5\% perturbation rate), while other datasets show similar trends. 
The less the performance of one architecture decreases, the more robust that architecture is.
~\cref{fig: intra} and~\cref{fig: inter} show boxplots for the robustness under various architectural designs.
For each subplot, the top figure displays the relationship between the operation choice and model test accuracy on the clean graph, while the bottom one shows the operation robustness via accuracy decreases. We make the following observations.

\xhdr{Intra Message-Passing Layer Design}
~\cref{fig: intra} shows the architecture's robustness for different intra-layer architecture designs. 
For simplicity, we only consider intra-layer operations for the first layer. We could see that architectural design plays a significant role in both architectural accuracy and robustness.
The use of pre-processing graph structure mask operations increases GNN models' robustness but sacrifices some accuracy. For the Cora dataset, \textit{NFS} is the most effective operation for pre-processing perturbed graph data.
\textit{GCN} is the most fragile correlation coefficient under Mettack. A plausible reason is that Mettack adopts a two-layer GCN model as the surrogate model to generate adversarial samples.
For neighbor aggregation operations, \textit{Sum} shows high accuracy and relatively robust performance. What's more, \textit{Max} is also able to generate some robust architectures.
Besides, combination functions are essential to enhance the robustness of GNNs as we could see a smaller accuracy decrease when using \textit{GIN} or \textit{SAGE} operations. 
Consequently, it is necessary to distinguish self- and neighbor-messages in the message-passing.

\xhdr{Inter Message-Passing Layer Design}
Our inter-layer designs include skip connections and layer aggregation operations. We also study how the number of message-passing layers affects the robustness of GNNs.
The results are summarized in~\cref{fig: inter}.
Skip connections help elevate both model accuracy and robustness. 
For the layer aggregation, there are no obvious differences, and \textit{LSTM} behaves slightly more robustly than the other two operations.
More interestingly, increasing the model depth makes GNNs more fragile under adversarial attacks. A 2-layer GNN shows reasonably good performance for the Cora dataset.

%% file: 10_conclusion.tex
\section{Conclusion and Limitations}
\label{sec:conclusion}

In this paper, we propose the first adversarially robust NAS framework for GNNs. 
We incorporate graph structure mask operations into the search space to enhance the defensive ability of GNNs. We also define a robustness metric that could effectively measure the architecture's robustness.
Experimental results demonstrate the effectiveness of our proposed approaches under adversarial attacks.
We empirically analyze the architectural robustness for different operations, which provides insights into the robustness mechanism behind GNN architectures.

Meanwhile, the lack of efficiency is a shared issue for many NAS methods. Since this weakness is not our main focus, we would like to leave this limitation as a future work. 
Societal impacts are discussed in the Appendix.

%% file: 12_appendix.tex
\appendix
\label{sec:appendix}


\onecolumn
\appendix
    \begin{center}
    \Large
    \textbf{Appendix: Adversarially Robust Neural Architecture Search \\for Graph Neural Networks}
     \\[20pt]
    \end{center}

\section{Broader Impact}
The key result of this research could aid the GNN family since G-RNA is able to search for robust architectures for learning graphs automatically. The proposed approach can simultaneously enhance the accuracy as well as the resilience to adversarial assaults of GNNs, allowing them to be used in more safety-critical applications including power grids, financial transactions, and transportation.

\section{Details of G-RNA}

\subsection{Detailed denotation for Graph Structure Masks
\label{app:denotation}}
Here, we explain the detailed denotation introduced in \cref{table: robust operations}.
For \textit{N}ode \textit{F}eature \textit{S}imilarity (\textit{NFS}),
\revision{$J_{ij}$ stands for the Jaccard similarity between node $i$ and $j$.}
For 
\textit{N}eighbor \textit{I}mportance \textit{E}stimation (\textit{NIE}), 
\revision{$\hat{\alpha}_{ij}^{(l)}$ is the importance weight as calculated in GNN-Guard~\cite{zhang2020gnnguard}.
For \textit{V}ariable \textit{P}ower \textit{O}perator (\textit{VPO}),
$V$ is a hyper-parameter indicating the power order of VPO.}

\subsection{Correlation coefficient operations\label{appendix: cor}}
For correlation coefficient operations, we follow the literature, and the detailed formulas are given in \cref{table: correlation coefficient}.

\subsection{Examples of popular GNNs based on search space of G-RNA}
Our search space could recover some classic and manually designed GNNs and state-of-the-art robust GNNs such as GCN-SVD, GCN-Jaccard, GNN-Guard, and VPN, as shown in \cref{table: GNNs}.

\begin{figure}[htbp]
	\centering
	\begin{minipage}{0.48\textwidth}
	    \captionof{table}{Commonly used correlation coefficient operations.
        \label{table: correlation coefficient}}
        \centering
        \resizebox{0.99\textwidth}{!}{%
        \begin{tabular}{l|l}
         \toprule
        $\mathcal{O}_{{e}}$ &  \textbf{Formula} \\
        \midrule
        \textit{Identity} & $e^{iden}_{ij}=1$\\
        \midrule
        \textit{GCN} &$e^{gcn}_{ij}=1/\sqrt{d_id_j}$ \\ 
        \midrule
        \textit{GAT}    &  $e^{gat}_{ij}={leaky\_relu}(\mathbf{W}_{l}\mathbf{h}_i + \mathbf{W}_{r}\mathbf{h}_j)$ \\
        \midrule
        \textit{GAT-Sym}   & $e^{sym}_{ij}=e^{gat}_{ij}+e^{gat}_{ji}$ \\
        \midrule
        \textit{Cos}   & $e^{cos}_{ij}=<\mathbf{W}_{l}\mathbf{h}_i, \mathbf{W}_{r}\mathbf{h}_j>$ \\
        \midrule
        \textit{Linear}   & $e^{lin}_{ij}=tanh(sum(\mathbf{W}_{l}\mathbf{h}_i))$ \\
        \midrule
        \textit{Gene-Linear}   & $e^{gene}_{ij}=\mathbf{W}_a tanh(\mathbf{W}_{l}\mathbf{h}_i + \mathbf{W}_{r}\mathbf{h}_j)$ \\
        \bottomrule
        \end{tabular}
        }
	\end{minipage}
\hfill
 \begin{minipage}{0.48\textwidth}
	    \captionof{table}{Recover existing GNN layers from our search space.\label{table: GNNs}}
        \centering
        \resizebox{0.99\textwidth}{!}{%
        \begin{tabular}{l|l}
        \toprule
        Method & $[\mathcal{O_D},\mathcal{O}_e, \mathcal{O}_{aggr},\mathcal{O}_{comb},\mathcal{O}_{skip}]$ \\
        \midrule
        GCN~\cite{ICLR2017SemiGCN} & [\textit{Identity, GCN, Sum, Identity, None}] \\
        \midrule
        JK-Net~\cite{Xu2018Representation} & [\textit{Identity, GCN, Sum, Identity, Skip}] \\
        \midrule
        GAT~\cite{velickovic2018gat} & [\textit{Identity, GAT, Sum, Identity, None}]\\ 
        \midrule
        GIN~\cite{xu2019how}   &  [\textit{Identity, Identity, Sum, GIN, None}] \\
        \midrule
        GraphSAGE~\cite{Hamilton2017Inductive} &  [\textit{Identity, Identity, Mean, SAGE, None}]\\
        \midrule
        GNN-Guard~\cite{zhang2020gnnguard} &  [\textit{NIE, GCN, Sum, Identity, None}]\\
        \midrule
        VPN~\cite{jin2021power} &  [\textit{VPO, GCN, Sum, Identity, None}]\\
        \bottomrule
        \end{tabular}
        }
	\end{minipage}
\end{figure}

\subsection{Evolutionary search algorithm \label{appendix: ea}}

Based on our proposed robustness metric, we adopt the evolutionary algorithm to search for the optimal robust architectures, as shown in \cref{alg: evolution}.
Inspired by the biological evolution process, the evolutionary algorithm solves optimization problems by mutation, crossover, and selection. 
The selection operation is conducted via a fitness function, which is set as $\operatorname{ACC}_{val}(\bm{\alpha}) + \lambda R(\bm{\alpha})$ in our algorithm. The inference function in Line 3 calculates the fitness score from the validation set.

\begin{algorithm}[htb]
\caption{Evolutionary Search Algorithm}
\label{alg: evolution}
\small{
\KwIn{The maximum iteration number $max\_iter$, supernet weights $\mathbf{W}$, the population size $P$, the mutation size $s$, the mutation probability $p$, the crossover size $n$, the original graph $G=(\mathbf{A},\mathbf{X})$, the perturbed adjacency matrices $\{\mathbf{A}'_t, t=1,..,T\}$, the number of optimal architectures $k$.} 
\KwOut{The top-K architectures with the highest robustness metric.}
\LinesNumbered
$candidates \gets \text{initialize\_population}(P)$ ;\\
\For {$iter=1,...,max\_iter$} {
 	$Q \gets \text{Inference}(candidates,G, \{\mathbf{A}'_i\});$ \\
 	$\text{Top-}k \gets \text{Select\_top}(Q,candidates,k)$;\\
 	$P_{\text{crossover}}\gets \text{Crossover}(\text{Top-}k, n)$;\\
 	$P_{\text{mutation}}\gets \text{Mutation}(\text{Top-}k, s, p)$;\\
 	$candidates \gets P_{\text{crossover}} \cup P_{\text{mutation}}$;
}
Return Top-$k$
}
\end{algorithm}

\section{Additional Experimental Results\label{addition_exp}}

\subsection{Defense Performance Against Targeted Attacks}
Targeted attacks and non-targeted attacks are two vital branches of the adversarial attack field. 
In this section, we compare our proposed methods to baselines under a targeted attack, Nettack~\cite{zugner2018adversarial}, as a complement to defensive results under non-targeted attacks. For each dataset, we choose 40 correctly classified nodes (10 nodes with the highest margin, 10 nodes with the lowest margin, and 20 nodes randomly) and report the average classification accuracy for these target nodes. Note that this setting leads to higher reported accuracy than the typical accuracy evaluated on the whole test set. We conduct this experiment for five runs and the defensive performance is shown in~\cref{tab: nettack}. It reads that G-RNA still outperforms the other baselines almost across all datasets under the perturbed setting while maintaining on-par performance on the clean graphs.

\begin{table}[htbp]
\centering
\caption{Defensive performance on targeted nodes (mean in percentages) under targeted attacks
Nettack. “-” indicates the result is unavailable due to the high time complexity of the model.}
\label{tab: nettack}
\resizebox{\columnwidth}{!}{
\begin{tabular}{llcccccccccc}
\toprule
Dataset & Model & GCN & GCN-JK & GAT & GAT-JK & GCN-Jaccard & Pro-GNN & GraphNAS & G-RNA \\ \midrule
\multirow{2}{*}{Cora} & No Attack & 93.25 & 94.75 & 94.75 & 92.25 & 98.50 & 97.50 & 96.00 & 94.95 \\
& \cellcolor[HTML]{BEBEBE} Attack & \cellcolor[HTML]{BEBEBE} 16.75 & \cellcolor[HTML]{BEBEBE} 16.25 & \cellcolor[HTML]{BEBEBE} 19.25 & \cellcolor[HTML]{BEBEBE} 23.75 & \cellcolor[HTML]{BEBEBE} 46.25 & \cellcolor[HTML]{BEBEBE} 50.00 & \cellcolor[HTML]{BEBEBE} 25.75 & \cellcolor[HTML]{87A96B} 51.25 \\\midrule
\multirow{2}{*}{Citeseer} & No Attack & 88.00 & 86.25 & 91.00 & 88.25 & 95.25 & 92.50 & 96.25 &  92.50\\
 & \cellcolor[HTML]{BEBEBE} Attack & \cellcolor[HTML]{BEBEBE} 14.75 & \cellcolor[HTML]{BEBEBE} 12.50 & \cellcolor[HTML]{BEBEBE} 17.25 & \cellcolor[HTML]{BEBEBE} 14.75 & \cellcolor[HTML]{BEBEBE} 22.00 & \cellcolor[HTML]{BEBEBE} 45.00 & \cellcolor[HTML]{BEBEBE} 9.00 & \cellcolor[HTML]{87A96B} 60.00 \\\midrule
\multirow{2}{*}{PubMed} & No Attack & 95.00 & 95.75 & 95.50 & 97.00 & 99.00 & - & 95.25 &  95.25\\
 & \cellcolor[HTML]{BEBEBE} Attack & \cellcolor[HTML]{BEBEBE} 12.50 & \cellcolor[HTML]{BEBEBE} 16.25 & \cellcolor[HTML]{BEBEBE} 13.00 & \cellcolor[HTML]{BEBEBE} 23.25 & \cellcolor[HTML]{BEBEBE} 1.50 & \cellcolor[HTML]{BEBEBE} - & \cellcolor[HTML]{BEBEBE} 9.00 & \cellcolor[HTML]{87A96B} 59.75  \\\midrule
\multirow{2}{*}{ogbn-arxiv} & No Attack & 85.00 & 90.50 & 92.50 & 93.50 & 95.00 & - & 97.50 &  92.25\\
 & \cellcolor[HTML]{BEBEBE} Attack & \cellcolor[HTML]{BEBEBE} 5.00 & \cellcolor[HTML]{BEBEBE} 0.50 & \cellcolor[HTML]{BEBEBE} 5.00 & \cellcolor[HTML]{BEBEBE} 6.75 & \cellcolor[HTML]{BEBEBE} 0.50 & \cellcolor[HTML]{BEBEBE} - & \cellcolor[HTML]{BEBEBE} 2.50 & \cellcolor[HTML]{87A96B} 22.50  \\\midrule
\multirow{2}{*}{Amazon\_photo} & No Attack & 96.25 & 98.25 & 89.50 & 98.50 & 99.75 & - & 97.50 & 94.50 \\
 & \cellcolor[HTML]{BEBEBE} Attack & \cellcolor[HTML]{BEBEBE} 12.50 & \cellcolor[HTML]{BEBEBE} 8.00 & \cellcolor[HTML]{BEBEBE} 12.00 & \cellcolor[HTML]{BEBEBE} 28.50 & \cellcolor[HTML]{BEBEBE} 9.00 & \cellcolor[HTML]{BEBEBE} - & \cellcolor[HTML]{BEBEBE} 5.00 & \cellcolor[HTML]{87A96B} 32.50  \\\bottomrule
\end{tabular}
}
\end{table}

\subsection{Defense on Heterophily Graph}\label{appendix: experiment-hete}
To better illustrate the effectiveness of our proposed method, we select one heterophily graph, Wisconsin and evaluate the performance of G-RNA on it. \cref{tab: heterophily} demonstrates the performance of various methods under both clean and perturbed data, where 5\% edges are manipulated by Mettack. We could find that even without the assumption of graph homophily, our method could outperform other baselines under both settings.  

\begin{table}[htbp]
\centering
\caption{Experimental result on heterophily graph.}
\label{tab: heterophily}
\resizebox{0.8\columnwidth}{!}{
\begin{tabular}{cccccc}
\toprule
\multicolumn{2}{c}{GCN} & \multicolumn{2}{c}{GAT} & \multicolumn{2}{c}{GCN-Jaccard} \\ 
\midrule
clean & perturbed &clean & perturbed &clean& perturbed \\
50.55±1.69 & 48.95±2.16 & 52.40±2.22 & 52.1±2.20 & 53.60±3.32 & 52.15±2.06 \\
\midrule
\multicolumn{2}{c}{Pro-GNN} & \multicolumn{2}{c}{GraphNAS} & \multicolumn{2}{c}{G-RNA} \\
 51.05±0.58&48.2±1.99  & 58.10±4.70 & 57.60±2.80 & 60.50±2.97 & 59.25±2.76 \\
 \bottomrule
\end{tabular}
}
\end{table}

\subsection{Details of Searched Architectures} \label{appendix: arch}
The optimal searched architectures by G-RNA for each dataset are listed in \cref{tab: arch}.
The optimal model depth searched is 2-layer for Cora, 3-layer for CiteSeer, and 1-layer for PubMed. 
All three architectures use \textit{NFS} as the pre-processing adjacency mask. \textit{NIE} is selected in the second layer for CiteSeer. 
Besides, \textit{LSTM} is chosen to be the layer aggregator for Cora and PubMed, while \textit{Concat} is selected for CiteSeer.

\begin{table}[htbp]
\centering
\caption{Searched architecture results.}
\label{tab: arch}
\resizebox{0.8\columnwidth}{!}{
\begin{tabular}{llcc}
 \toprule
Dataset & Layer & Intra-layer Operations & Layer aggregator \\ 
\midrule
\multirow{2}{*}{Cora} & layer1: \textit{Skip} & {[}\textit{NFS, GAT-Cos, Sum, Identity}{]} & \multirow{2}{*}{\textit{LSTM}} \\
 & layer2: \textit{Skip} & {[}\textit{Identity, Identity, Sum, Identity}{]} &  \\ 
 \midrule
\multirow{3}{*}{CiteSeer} & 
layer1: \textit{Skip} & {[}\textit{NFS, Gene-Linear, Mean, Identity}{]} & \multirow{3}{*}{\textit{Concat}} \\
 & 
 layer2: \textit{None} & {[}\textit{NIE, GAT, Max, GIN}{]} \\
 &
 layer3: \textit{Skip} &{[}\textit{VPO, Identity, Max, Identity}{]} 
 \\ 
 \midrule
 PubMed & layer1: \textit{Skip} & [\textit{NFS, Identity, Max, SAGE\textit}]
 & \textit{LSTM}\\
  \midrule
 \multirow{2}{*}{ogbn-arxiv} & layer1: \textit{Skip} & {[}\textit{NFS, gat, Sum, GIN}{]} & \multirow{2}{*}{\textit{Concat}} \\
 & layer2: \textit{Skip} & {[}\textit{Identity, GAT-Sym, Sum, GIN}{]} &  \\
  \midrule
 \multirow{3}{*}{Amazon\_{photo}} & layer1: \textit{Skip} & [\textit{NFS, GAT-Linear, Sum, GIN}] & \multirow{3}{*}{\textit{Concat}} \\
 & layer2: \textit{Skip} & [\textit{Identity, Identity, Sum, GIN}] &  \\
 & layer3: \textit{Skip} & [\textit{Identity, GAT-Sym, Sum, Identity}] &  \\
 \bottomrule
\end{tabular}
}
\end{table}

\subsection{Sensitivity Analysis}\label{appendix: ablation}
\begin{figure}[htbp]
	\centering
	\begin{minipage}{0.48\textwidth}
		\centering
		\includegraphics[width=\textwidth]{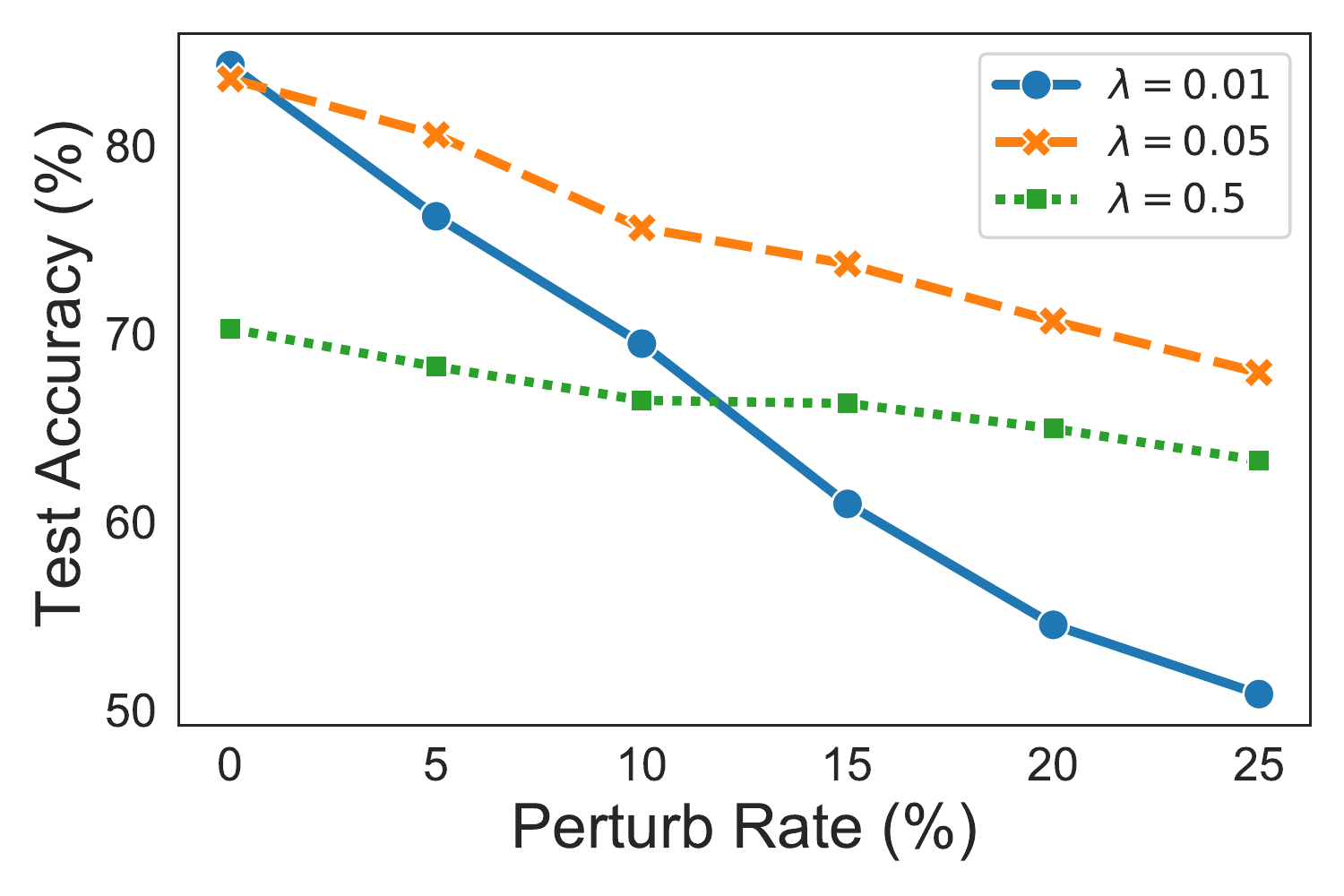}
		\caption{Parameter sensitivity analysis for $\lambda$.\label{fig: ablation}}
	\end{minipage}
	\hfill
	\begin{minipage}{0.38\textwidth}
	    \captionof{table}{Running time ($hours$) comparison on Cora dataset.\label{tab:timeCost}}
        \centering
        \begin{tabular}{lll}
        \toprule
        Pro-GNN & GraphNAS & G-RNA\\
        \midrule
        0.22 & 4.12 & 2.23 \\
        \bottomrule
        \end{tabular}
	\end{minipage}
\end{figure}

To in-depth understand the effect of the hyper-parameter $\lambda$ in the search process, we conduct an ablation study on its sensitivity.
\cref{fig: ablation} shows the performance for $\lambda={0.01, 0.05, 0.5}$ on Cora dataset. The experiments on the other datasets show similar patterns.
When $\lambda$ is very small, e.g., $0.01$, the performance on clean data is gratifying, but the performance drops fast under adversarial perturbations.
On the contrary, when $\lambda$ is relatively large, e.g., $0.5$, the prediction accuracy on the clean graph is poor, but the performance is stable even when the graph is heavily attacked.
As a result, we need to properly balance the accuracy and robustness tradeoff by choosing a suitable $\lambda$. In practice, we find that tuning $\lambda$ in the range of $[0.03,0.3]$ usually leads to satisfactory results.

\subsection{Computational Efficiency Analysis}
We also empirically demonstrate the computational efficiency of G-RNA in comparison with two comprehensive baselines, GraphNAS and Pro-GNN, on the Cora dataset in \cref{tab:timeCost}.
Due to the usage of graph structure mask operations and the evolutionary algorithm, our methods are not as efficient as darts-based NAS methods or plain robust GNN solutions. However, we can find that the computation cost of our G-RNA is comparable to GraphNAS. This lack of efficiency is consistent with one of the limitations that we claimed in \cref{sec:conclusion}, which is a shared issue for many NAS methods and we choose to leave its improvement as a future work.

\section{Experimental Setup}\label{appendix: Experiment Details}
In this section, we start with the dataset statistics and then describe the implementing details for G-RNA and other baselines.

\subsection{Dataset \label{appendix: dataset}}
We mainly consider the semi-supervised node classification task on three citation graph datasets~\cite{Dataset2008Citeseer}\footnote{https://github.com/kimiyoung/planetoid/tree/master/data}, including Cora, CiteSeer, and PubMed. For each graph, we randomly select 10\% nodes for training, 10\% nodes for validation, and the rest 80\% nodes for testing to be consistent with existing literature. 
Besides, we also evaluate the effectiveness of G-RNA on one heterophily graph, one citation network from Open Graph Benchmark (OGB)~\cite{hu2020open}, and a co-purchase network that intends to predict the product category on Amazon~\cite{mcauley2015image}. The specific statistics for each dataset are shown in \cref{tab:dataset}.

\begin{table}[htbp]
\centering
\caption{Dataset statistics.}
\label{tab:dataset}
\begin{tabular}{lccccc}
\toprule
\textbf{Dataset} & \textbf{\# Nodes} & \textbf{\# Edges} & \textbf{\# Features} &  \textbf{\# Classes} \\ \midrule
Cora & 2,708 & 5,429 & 1,433  & 7 \\
Citeseer & 3,327 & 4,732 & 3,703  & 6 \\
PubMed & 19,717 & 44,338 & 500 & 3 \\ 
ogbn-arxiv & 50,802 & 108,554 & 128 & 40\\
Amazon\_photo &7,650 & 119,081 &  745 & 8 \\
Winsconsin & 251 & 499 & 1703 & 5\\

\bottomrule
\end{tabular}
\end{table}

\subsection{Baselines \label{appendix: baselines}}
In order to validate the effectiveness and robustness of our G-RNA, we compare it with state-of-the-art GNNs, manually-designed robust GNNs, and Graph NAS methods: 
\begin{itemize}[leftmargin = 0.5cm]
    \item \textbf{GCN}~\cite{ICLR2017SemiGCN}: Graph Convolution Network (GCN) is the pioneer of GNN and represents as a classic victim model against adversarial attacks.
    \item \textbf{GCN-JK}~\cite{Xu2018Representation}: GCN-JK combines GCN with jumping knowledge networks (JK-Net). JK-Net adds skip connections among different hidden layers and adaptively learns node-wise neighborhoods when aggregating node representations.
    \item \textbf{GAT}~\cite{velickovic2018gat}: Graph Attention Network (GAT) utilizes the self-attention mechanism to learn different weights for edges.
    \item \textbf{GAT-JK}~\cite{Xu2018Representation}: GAT-JK is the combination of GAT and JK-Net. Again, We use GNNs and their JK-Net variants to study the effect of the JK-Net backbone.
    \item \textbf{RGCN}~\cite{RGCN}: Robust Graph Convolutional Network (RGCN) is an attention-based defense model by treating node representations as Gaussian distribution and assigning less attention to nodes with high variance. 
    \item \textbf{GCN-Jaccard}~\cite{wu2019adversarial}: GCN-Jaccard conducts edge pruning according to the jaccard similarity among node representations. It is a simple yet effective pre-processing defensive baseline.
    \item \textbf{Pro-GNN}~\cite{jin2020graph}: Based on the graph properties of sparsity, low rank, and feature smoothness, Property GNN (Pro-GNN) learns parameters and purifies the adjacency matrix at the same time. 
    \item \textbf{DropEdge}~\cite{rong2019dropedge}: DropEdge randomly drops edges for each graph convolution layer.
    \item \textbf{PTDNet}\footnote{https://github.com/flyingdoog/PTDNet}~\cite{luo2021learning}: PTDNet improves the robustness of GNNs and prunes task-irrelevant edges with parameterized networks. 
    \item \textbf{GraphNAS}\footnote{https://github.com/GraphNAS/GraphNAS}~\cite{ijcai2020-195}: GraphNAS serves as a baseline that validates the robustness of plain graph NAS under adversarial attacks.
    \item \revision{\textbf{GASSO}}\footnote{https://github.com/THUMNLab/AutoGL}~\cite{qin2021graph}: GASSO conducts graph structure learning with architecture search to search under potential noisy graph data.
\end{itemize}
Except for GraphNAS, we use the public implementation for all baselines via PyTorch Geometric (PyG)~\cite{Fey/Lenssen/2019} and DeepRobust~\cite{li2020deeprobust}.

\subsection{Parameter Settings}
To search for the optimal robust GNNs, we first construct a supernet and train it afterward. Then, we search with our robustness metric using the weights of the trained supernet. At last, we retrain the top-5 selected architectures from scratch.
For the construction of the supernet, we limit the maximum layer number to $3$.

Since the random attack is one simple yet effective attack method, we use random attack as our attack proxy and generate 5 perturbed graphs ($T=5$) with a 5\% perturbation rate. Meanwhile, it is easy to implement, so that we use it to evaluate the robustness of architectures without the knowledge of the specific attacker. Some other attack algorithms could also be utilized to generate adversarial examples for the evaluation of the robustness metric if some prior knowledge from the attacker is given.

For the proposed G-RNA, the supernet is trained for 1,000 epochs with a learning rate of 0.005 and a weight decay of 3e-4. The linear dropout rate is fixed at 0.5, and the attention dropout rate is fixed at 0.6. 
$\lambda$ is determined through grid-search and selected as 0.05 for Cora and PubMed, and 0.1 for CiteSeer. For robust operations, the reconstruction rank $r$ for LRA is set to 20, and the threshold for NFS operation $\gamma=0.01$. We maintain the power order $V$ of VPO to be 2 for all datasets.
In the generic search algorithm, we set population size $P=50$, mutation size $s=25$, mutation probability $p=0.1$, crossover size $n=25$, and optimal architecture number $k=10$.
After we finish the search, we continue to tune the hyper-parameters using hyperopt\footnote{https://github.com/hyperopt/hyperopt} with the following options to gain the best results:
\begin{itemize}[leftmargin = 0.5cm]
    \item hidden size: \{16, 32, 64, 128, 256\}
    \item learning rate: \{0.005, 0.01\}
    \item weight decay: \{5e-3, 1e-3, 5e-4, 1e-4\}
    \item optimizer: \{adam, adagrad\}
    \item linear dropout: \{0, 0.3, 0.5, 0.7\}
    \item attention dropout: \{0.5, 0.6, 0.7\}
\end{itemize}

For all models, we train them on ogbn-arxiv for 500 epochs and on all the other datasets for 200 epochs.
For vanilla GNNs, we set the learning rate as 0.005 using the Adam optimizer. Other hyper-parameters are kept the same as the original papers.
For GNN-Jaccard, we tune the threshold for Jaccard similarly from \{0.01,0.02,0.03,0.04,0.05\}.
For RGCN, the hidden size is tuned from \{16, 32, 64, 128\}.
For Pro-GNN, we follow the original settings to fix the hidden size as 64 for Cora and CiteSeer. 
The results of Pro-GNN for the PubMed dataset are not available due to its high time complexity. In addition, its performances differ from the original paper due to the data split and whether to select the largest connected component.
For GraphNAS, we keep the same setting as reported in the original paper and rerun it because their code leaks the test set of data.
We ran all experiments on a single machine with a 16GB GeForce GTX TITAN X GPU.

\section{Extension to feature attack}
\revision{To extend our method for feature-based attacks}, we redefine the robustness metric as 
\begin{equation}
    -\mathbb{E}_{\mathbf{X}'} \left[\frac{1}{N}\sum_{i=1}^N D_{KL}\left( f(\mathbf{X})_i||f(\mathbf{X}')_i\right) \right], \\ 
\mathbf{X}' = \mathcal{T}_{\Delta}(\mathbf{X}).
\end{equation}

We leave the validation of the experimental effectiveness for G-RNA's feature attack extension as our future work.